\documentclass[sigconf]{acmart}

\AtBeginDocument{%
  }

\setcopyright{acmlicensed}
\copyrightyear{2018}
\acmYear{2018}
\acmDOI{XXXXXXX.XXXXXXX}

\acmJournal{TOG}
\acmVolume{37}
\acmNumber{4}
\acmArticle{111}
\acmMonth{8}

\usepackage{natbib}
\setcitestyle{numbers,square}
\usepackage{amsmath}
\usepackage{adjustbox}
\usepackage{enumitem}
\usepackage{textcomp}
\usepackage{gensymb}
\usepackage{multirow}
\usepackage{array}       
\usepackage{graphicx}    

\usepackage{booktabs}    
\usepackage{caption}     
\usepackage{float}
\usepackage{tabularx}
\usepackage{tabularray}
\usepackage{makecell} 
\usepackage{stfloats}
\makeatletter
\def\@mkauthorsaddresses{}
\makeatother
\begin{document}

\title{Differentiable Fast Top-K Selection for Large-Scale Recommendation}

\author{Yanjie Zhu$^*$}
\affiliation{
 \institution{Xi'an Jiaotong University}
 \city{Xi'an}
 \country{China}
}
\email{zhuyanjie@stu.xjtu.edu.cn}

\author{Zhen Zhang$^*$}
\affiliation{
 \institution{Kuaishou Technology}
 \city{Beijing}
 \country{China}
}
\thanks{*These authors contributed equally. The work was done during Yanjie Zhu's internship at Kuaishou. $\dag$ Corresponding author.}
\email{zhangzhen24@kuaishou.com}

\author{Yunli Wang$^*$}
\affiliation{
 \institution{Kuaishou Technology}
 \city{Beijing}
 \country{China}
}
\email{wangyunli@kuaishou.com}

\author{Zhiqiang Wang}
\affiliation{
 \institution{Kuaishou Technology}
 \city{Beijing}
 \country{China}
}
\email{wangzhiqiang03@kuaishou.com}

\author{Yu Li}
\affiliation{
 \institution{Kuaishou Technology}
 \city{Beijing}
 \country{China}
}
\email{liyu26@kuaishou.com}

\author{Rufan Zhou}
\affiliation{
 \institution{Kuaishou Technology}
 \city{Beijing}
 \country{China}
}
\email{zhourufan03@kuaishou.com}

\author{Shiyang Wen}
\affiliation{
 \institution{Kuaishou Technology}
 \city{Beijing}
 \country{China}
}
\email{wenshiyang@kuaishou.com}

\author{Peng Jiang}
\affiliation{
 \institution{Kuaishou Technology}
 \city{Beijing}
 \country{China}
}
\email{jiangpeng@kuaishou.com}

\author{Chenhao Lin$^\dag$}
\affiliation{
 \institution{Xi'an Jiaotong University}
 \city{Xi'an}
 \country{China}
}
\email{linchenhao@xjtu.edu.cn}

\author{Jian Yang$^\dag$}
\affiliation{
 \institution{M-A-P}
 \city{Beijing}
 \country{China}
}
\email{jiaya@map}
\renewcommand{\shortauthors}{Yanjie Zhu, Zhen Zhang, Yunli Wang et al.}

\begin{abstract}
Cascade ranking is a widely adopted paradigm in large-scale information retrieval systems for Top-K item selection. However, the Top-K operator is non-differentiable, hindering end-to-end training. Existing methods include Learning-to-Rank approaches (e.g., LambdaLoss), which optimize ranking metrics like NDCG and suffer from objective misalignment, and differentiable sorting-based methods (e.g., ARF, LCRON), which relax permutation matrices for direct Top-K optimization but introduce gradient conflicts through matrix aggregation.
A promising alternative is to directly construct a differentiable approximation of the Top-K selection operator, bypassing the use of soft permutation matrices. However, even state-of-the-art differentiable Top-K operator (e.g., LapSum) require $O(n \log n)$ complexity due to their dependence on sorting for solving the threshold.
Thus, we propose DFTopK, a novel differentiable Top-K operator achieving optimal $O(n)$ time complexity. By relaxing normalization constraints, DFTopK admits a closed-form solution and avoids sorting. DFTopK also avoids the gradient conflicts inherent in differentiable sorting-based methods.
We evaluate DFTopK on both the public benchmark RecFLow and an industrial system. Experimental results show that DFTopK significantly improves training efficiency while achieving superior performance, which enables us to scale up training samples more efficiently. In the online A/B test, DFTopK yielded a +1.77\% revenue lift with the same computational budget compared to the baseline. 
To the best of our knowledge, this work is the first to introduce differentiable Top-K operators into recommendation systems and the first to propose a closed-form differentiable Top-K operator with linear-time complexity.
We have open-sourced our implementation to facilitate future research in both academia and industry.

\end{abstract}



\received{20 February 2007}
\received[revised]{12 March 2009}
\received[accepted]{5 June 2009}

\maketitle

\begin{figure}[H]
\centering
\includegraphics[
    width=\linewidth,
    trim=0 21cm 0 1.3cm,
    clip
]{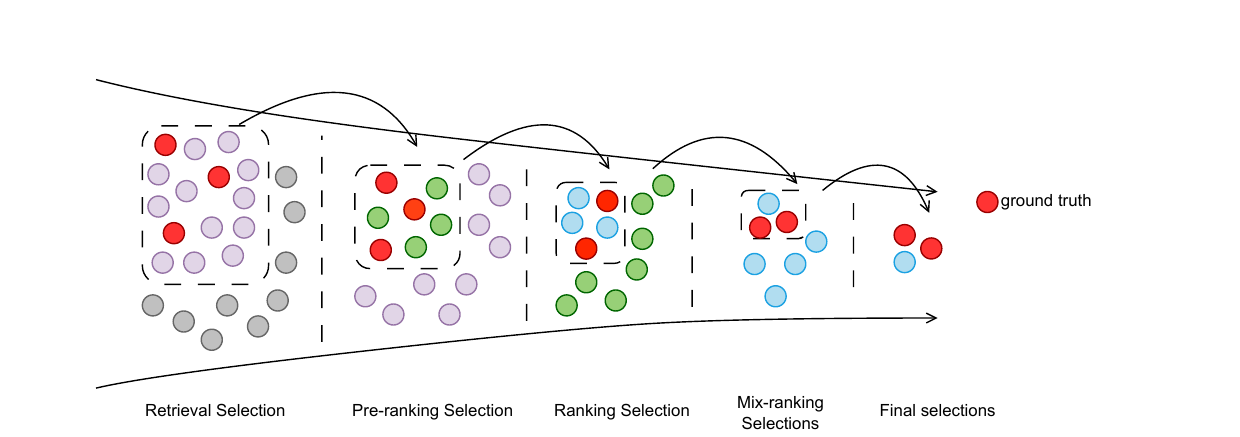}
\caption{A typical cascade ranking architecture. Including four stages: Matching, Pre-ranking, Ranking, and Mix-ranking. The red points represent the ground truth for the selection.}
\label{fig:cascade}
\end{figure}

\section{Introduction}
\label{intro}
Cascade ranking~\cite{covington2016deep,qin2022rankflow,wang2011cascade,zheng2024full,liu2017cascade,scalingAdRetri2024} is the standard architecture in large-scale recommendation and advertising, supporting the core operations of major industrial platforms. This multistage funnel skillfully balances computational cost with the quality of personalized selections. It works by progressively filtering billions of items through stages like Matching, Pre-ranking, and Ranking to yield a small set of optimal results. 
As illustrated in Figure ~\ref{fig:cascade}, the objective is to select ground truth items (referred to as the red points in Figure ~\ref{fig:cascade}) as the final outputs.
Central to each stage is the recurring task of Top-K selection. This operation is fundamental to the entire workflow, from retrieving the initial Top-K candidates to presenting the final user-facing recommendations. Although Top‑K selection is widely used, it is inherently non‑differentiable due to its discrete and piecewise‑constant nature, making direct end‑to‑end optimization in deep learning challenging.

To bridge the gap left by non-differentiable operators, traditional Learning-to-Rank (LTR) methods~\cite{burges2005learning,cao2006adapting,jagerman2022optimizing,li2007mcrank,taylor2008softrank,wang2018lambdaloss,wu2009smoothing,xu2007adarank,zheng2007general} employ pairwise and listwise losses for indirect optimization. For example, RankNet~\cite{burges2010ranknet} focuses on the relative ordering of item pairs, while listwise approaches such as LambdaLoss~\cite{wang2018lambdaloss} are designed to optimize full-list ranking metrics like NDCG.  However, these methods suffer from objective misalignment with the Top-K problem, as their primary goal is to learn the relative order of items, whereas the core requirement of Top-K selection is simply to identify the correct set of top items.

\begin{figure*}[!t]
  \centering
  \includegraphics[width=0.8\textwidth]{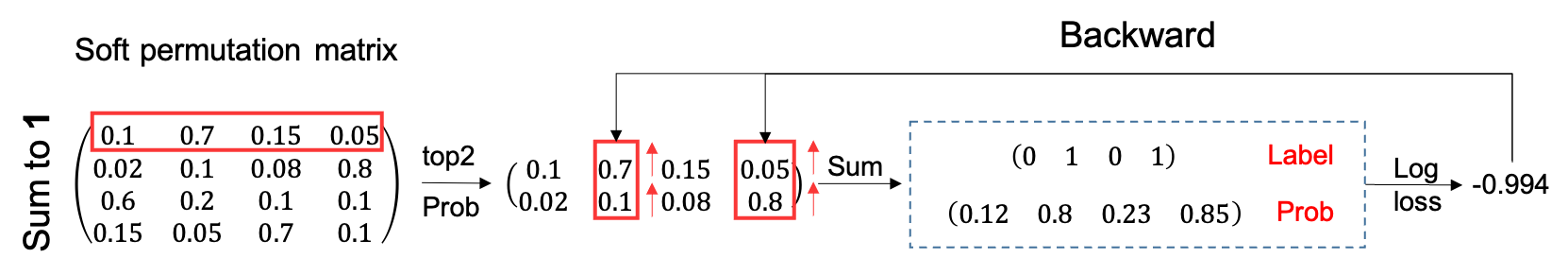}
    \vspace{-8pt}
    \caption{Gradient conflict in soft permutation matrix. In NeuralSort, the sum-to-one constraint in each row inevitably induces zero-sum competition among ground-truth items, causing gradient conflict in every row.}
  \label{fig:grad_conflict}
\end{figure*}

To address this, methods based on differentiable permutation matrices~\cite{grover2019stochastic, prillo2020softsort, petersen2022monotonic} have emerged, such as ARF~\cite{wang2024adaptive} and LCRON~\cite{wang2025learning}, which model the probability of the top-$k$ set via soft permutation matrices.
This enables the direct E2E optimization of metrics like Set Recall, offering a tighter alignment with the Top-K objective than traditional LTR methods. 
However, using differentiable permutation matrices for Top-K modeling, which involves aggregating the Top-K rows, presents two critical issues: 1) The sum-to-one constraint on each rank's probability distribution inevitably causes gradient conflicts among ground-truth items, as illustrated in Figure~\ref{fig:grad_conflict}. In other words, these methods would like the probability of ground-truth items being ranked in the top-k to increase simultaneously, but clearly, a single item cannot be ranked both at position $k$, $k-1$, $k-2$, $...$, and $1$ at the same time. 2) The method's $O(n^2)$ time complexity is a major practical bottleneck, as it often necessitates reducing the training sample size, thereby limiting model performance.

To address these issues, a more intuitive approach is to directly model the Top-K selection process itself. Recently, LapSum~\cite{struski2025lapsum} implements soft top-k selection through threshold comparison, while Sparse Top-K~\cite{sander2023fast} combines polytope sorting with isotonic optimization to maintain sparsity and differentiability. 
By bypassing the use of differentiable permutation matrices, these methods inherently avoid the gradient conflicts caused by the sum-to-one constraints at each rank position.
Although these operators effectively reduce gradient conflicts and improve training efficiency, their computational complexity remains a bottleneck for large-scale tasks. Their reliance on sorting or iterative solvers limits their best-case time complexity to $O(n \log n)$, which is algorithmically suboptimal.
Moreover, as discussed in LapSum~\cite{struski2025lapsum}, there also exist some other strategies for modeling Top-K selection, such as smooth approximations~\cite{berrada2018smooth, garcin2022stochastic} and optimization-based approaches~\cite{xie2020differentiable}. However, these methods suffer from either introducing large approximation errors or incurring high computational costs.

\textbf{To sum up, no existing method adequately addresses all of the aforementioned challenges, highlighting the need for a more efficient and comprehensive Top-K operator for recommendation systems.}

To address these challenges, we introduce DFTopK, a novel operator that reformulates the Top-K selection problem. The core innovation of our approach lies in relaxing the strict normalization constraints typically imposed by strict Top-K methods. By formulating a controllable approximation to the sum-to-K probability distribution for each item, DFTopK weakens the coupling among items in the optimization process. 
This reformulation not only reduces the computational complexity of the operator to linear $O(n)$ but also substantially mitigates the gradient conflicts that arise from the competitive nature of full ranking.
Additionally, a systematic theoretical analysis of the DFTopK operator, supported by a rigorous derivation of its gradient formulation, validates its inherent advantages.

To validate the effectiveness of our proposed method, we conduct comprehensive experiments on the RecFlow~\cite{liu2024recflow} benchmark under the LCRON training framework~\cite{wang2025learning}. Within this benchmark, we replace the original NeuralSort~\cite{grover2019stochastic} operator in LCRON with our proposed DFTopK operator, alongside several baselines based on permutation matrices and other Top-K operators. Experimental results under a streaming evaluation protocol simulating continuous online learning demonstrate that our method achieves highly competitive performance. We further compare the per-step forward and backward computation time on both GPU and CPU, where our method consistently achieves optimal efficiency. This demonstrates its effectiveness and efficiency.  
Furthermore, we perform an online A/B test in a real-world advertising system. DFTopK, under the same training sample conditions, achieves a 0.57\% increase in advertising revenue and a 0.9\% improvement in advertising conversions compared to the baseline, while reducing the average rungraph time per impression by 15.3\%. When the training data scale is increased, the advantages of DFTopK become even more pronounced, with a 1.77\% increase in advertising revenue and a 1.5\% improvement in advertising conversions, demonstrating significant practical value.

To the best of our knowledge, our main contributions are threefold: 1) We make the first attempt to introduce a differentiable Top-K operator into industrial recommendation systems, validating the feasibility and potential of this technical direction. 2) We propose the first closed-form differentiable Top-K operator with linear complexity, which replaces sorting operations through a controlled approximation. This approach not only improves computational efficiency but also effectively mitigates the inherent gradient conflicts in differentiable sorting methods, thereby ensuring training stability and enhancing model performance. 3) We have demonstrated the superior effectiveness and efficiency of DFTopK through extensive offline experiments. Furthermore, the method has been deployed in a real-world advertising system to investigate its practical impact in real-world applications.

\section{Related Work}
Current approaches for modeling the Top-K problem in recommendation systems predominantly fall into two major categories: permutation-based methods and differentiable Top-K operators. Given their prevalence and relevance to our work, this section will primarily focus on reviewing these two families of solutions.

\noindent\textbf{Traditional LTR.} Before differentiable ranking operators, Top-K problems in recommendation systems were mainly addressed through LTR methods. These approaches employ surrogate losses, typically categorized into pairwise and listwise formulations.
Pairwise methods, such as RankNet \cite{burges2005learning}, learn relative preferences between item pairs and are computationally efficient, but they only optimize local orderings and often misalign with global Top-K metrics like Recall and NDCG. Listwise methods, including ListNet \cite{cao2007learning} and ListMLE \cite{xia2008listwise}, optimize over entire lists, offering a closer alignment with evaluation metrics. Extensions such as LambdaRank and LambdaMART \cite{burges2010ranknet} further improved their practical utility by designing gradients consistent with ranking measures.
Despite their effectiveness, both pairwise and listwise approaches optimize for the full ranking, which imposes stronger conditions than required for Top-K selection. This misalignment between surrogate objectives and actual task goals \cite{bruch2021alternative} has motivated the exploration of permutation-based and differentiable Top-K operators for more direct and efficient modeling.

\noindent\textbf{Differentiable Sorting Operator.} A prominent line of research for differentiable ranking involves learning continuous relaxations of permutations. These methods~\cite{prillo2020softsort, grover2019stochastic, petersen2022monotonic} aim to construct a soft permutation matrix $P$, where $P_{ij}$ represents the probability that item $i$ is in rank $j$. This formulation allows for the end-to-end optimization of listwise metrics like NDCG and Recall by framing the ranking problem in a probabilistic manner.

NeuralSort~\cite{grover2019stochastic} is a pioneering approach which generates the permutation matrix by performing pairwise comparisons across all items, but this results in a computationally intensive $O\left(n^2\right)$ complexity. The smoothness of its output is controlled by a temperature parameter $\tau$, which requires careful tuning to balance approximation quality and gradient flow. Furthermore, by deriving a reparameterized gradient estimator for the classic Plackett-Luce probabilistic model, NeuralSort extends its capability from deterministic sorting to learning within the vast, stochastic space of permutations.

To improve scalability, SoftSort~\cite{prillo2020softsort} was introduced as a more efficient $O\left(n^2\right)$ alternative. Its core insight is to compute the permutation by measuring the pairwise distances between the input scores and their sorted counterparts, effectively projecting the scores onto an ideal ranking. This reliance on standard sorting algorithms makes it significantly faster for longer sequences.

A distinct paradigm is offered by DiffSort~\cite{petersen2022monotonic}, which makes classical sorting networks (odd-even or bitonic networks) differentiable. It achieves this by replacing the discrete min/max functions in the network's comparators with a smooth, probabilistic interpolation. This provides a more structured and localized approximation to sorting with a favorable $O\left(n^2(\log n)^2\right)$ complexity, offering a different trade-off between fidelity and gradient stability.

\noindent\textbf{Differentiable Top-K Operator.}
Beyond permutation-based approaches, another significant line of research tackles the nondifferentiability of Top-K selection by reframing it as a continuous optimization problem, often providing stronger theoretical guarantees and superior efficiency.

LapSum~\cite{struski2025lapsum} is a novel, unified framework for differentiable ranking, sorting and Top-K selection, grounded in the theory of the sum of Laplace distributions. It successfully addresses a series of soft ordering problems, including soft sorting and soft Top-K, by providing a single, theoretically rigorous, closed-form solution. Compared to existing methods, LapSum dramatically reduces computational and memory complexity—achieving $O\left(n \log n\right)$ time and $O(n)$ memory—while maintaining state-of-the-art (SOTA) performance. This resolves long-standing efficiency and scalability bottlenecks in the field, offering a powerful, practical, and efficient solution for large-scale, differentiable ranking problems.

Sander et al.~\cite{sander2023fast}introduced a significant breakthrough in achieving sparsity for differentiable Top-K operators from a convex analysis perspective. Their method formulates the operator as a p-norm regularized program over the permutahedron, which they show can be reduced to an isotonic optimization problem. This approach uniquely yields an operator that is simultaneously sparse and fully differentiable. Algorithmically, it is highly efficient and solvable via PAV or Dykstra's algorithm with an $O\left(n\log n\right)$ time complexity. 

Despite their innovations, the computational overhead and gradient stability of these methods remain key challenges, motivating the search for more efficient operators.
\section{Formulation of the Top-K Set Selection Problem}
\label{sec:problem formulation}
We begin by formulating the Top-K set selection problem, which is central to our work.
The core objective in many recommendation and ranking scenarios can be formalized as a Top-K set selection problem. Let 
 $\mathcal{I}=\{i_{1}, i_{2},\cdots,i_{N}\}$ be the set of all candidate items. Given an input context (e.g., a user profile), a model produces a score vector $x\in\mathbb{R}^N$, where $x_{i}$is the predicted score for item $i$. Let $\mathbb{G}\subset\mathbb{I}$ be the set of ground truth items, with a cardinality of $|\mathbb{G}|=K$, where $K\geqslant1$.
The model's task is to generate a ranked list of items. We denote the set of indices of the $Top-M$ items predicted by the model as 
$X_{M}(x)$, where $M\geqslant K$. The ideal outcome is to have all ground truth items appear in this predicted $Top-M$. Formally, the optimization goal is to maximize the overlap between the ground truth set and the predicted set:
$$
maximize \quad |\mathbb{G}\cap X_{M}(x)|
$$
This objective is equivalent to maximizing the $Recall@K@M$ for the $K$ ground truth items. In the subsequent sections, we will discuss how to construct a differentiable surrogate for this discrete, set-based objective.

\section{Methodology}
This section delineates our approach to modeling the Top-K problem using differentiable operators. We begin by analyzing two predominant paradigms: differentiable sorting operators and differentiable Top-K operators. We first demonstrate that the sorting-based approach inherently suffers from gradient conflicts due to its competitive nature. We then establish the desirable properties for a differentiable Top-K operator and discuss how existing methods, despite satisfying these properties, remain computationally inefficient for large-scale applications. 
To address this, we propose DFTopK, which preserves the essential characteristics of a strict differentiable operator while allowing the constraint $\sum_{i=1}^{N} f_{k}(x_{i})=k$ to be approximately satisfied, with the approximation quality controlled by a temperature parameter. This enables an $O(n)$ linear-time complexity.
\textbf{This linear complexity is particularly important for industrial applications, as it enables the use of a significantly larger number of samples under the same computational budget, leading to more robust and effective models.} Moreover, we conduct a rigorous gradient analysis of the operator and its corresponding loss function, demonstrating that it preserves favorable gradient properties during training.

\subsection{Gradient Conflict in Differentiable Sorting Operator}
We begin by introducing differentiable sorting operators to model the Top-K problem and elucidating the root causes of the gradient conflict issue. In the discrete domain, the result of sorting a sequence of $N$ items, let us denote it by $x$, is uniquely represented by a hard permutation matrix $P$. This is a square binary matrix of size $N\times N$, where $P_{ij}=1$ if $x_j$ is at rank $i$ and 0 otherwise. Formally, such a matrix contains exactly one entry of 1 in each row and column, ensuring a one-to-one mapping between items and ranks.

\noindent\textbf{Example:} Let input vector $x=\left[4,1,3,2\right]^T$, the permutation matrix is $P = \left[\begin{smallmatrix}
    1 & 0 & 0 & 0 \\
    0 & 0 & 1 & 0 \\
    0 & 0 & 0 & 1 \\
    0 & 1 & 0 & 0
\end{smallmatrix}\right]$.
Sorted vector $x_{sorted}=Px = \left[\begin{smallmatrix}
    1 & 0 & 0 & 0 \\
    0 & 0 & 1 & 0 \\
    0 & 0 & 0 & 1 \\
    0 & 1 & 0 & 0
\end{smallmatrix}\right] \cdot [4,1,3,2]^{T} = [4,3,2,1]^{T} $.

However, the discrete and piecewise constant nature of hard permutation matrices renders them non-differentiable. To address this limitation, a significant research direction has focused on developing continuous relaxations through soft permutation matrices. The differentiable permutation matrix corresponding to the input sequence $x$ can be obtained through a specialized differentiable continuous function $\hat{P}=\Gamma(x)$. This soft permutation matrix maintains the $N \times N$ dimensionality, but its elements $\hat{P}_{i,j}$ represent continuous values indicating the probability or likelihood of item $j$ being assigned to rank $i$. 
The Top-K objective is defined via a soft permutation matrix to compute the expected number of relevant items within the Top-K positions.
$$\mathbb{E}\left\lbrack TopK\right\rbrack=\sum\limits_{j=1}^{N}y_{j}\left(\sum\limits_{i=1}^{K} \hat{P}_{i,j}\right)$$
where $y$ denotes the ground truth. During the optimization process, while we aim to increase the values of $\sum_{i=1}^{K} \hat{P}_{i,topk}$
 corresponding to Top-K elements, the inherent properties of the sorting matrix (which satisfy $\sum_{j=1}^{N} \hat{P}_{i,j}=1$) introduce competition among elements, consequently leading to gradient conflicts. \textbf{In other words, while we aim to increase the probability of the ground truth at all Top-K positions, the constraint that each Top-K position can only be assigned to one element inevitably leads to gradient conflicts.}

\subsection{Properties of Differentiable Top-K Operators and Limitations of Existing Methods}
A more direct and simpler approach to modeling the Top-K problem is to construct a differentiable Top-K operator. In the discrete domain, the standard hard Top-K operator $F_k(x)$ acts on an input score vector $x$ and returns a new vector $x_{\text{topk}}$ where the top K values of $x$ are preserved and all other entries are set to zero. A hard mask $M$ can uniquely represent this selection process. This mask, $M\in \{0, 1\}_{N}$, with its entries $m_i$ being 1 for the top $K$ indices and 0 for others.

\noindent\textbf{Example:} Let input vector $x=[4,1,3,2]^{T}$, the top-2 mask is $M = [1,0,1,0]^{T}$. Top-K vector $x_{topk} = x \odot M = [4,0,3,0]^{T}  $.

This fundamental operation is non-differentiable because of its abrupt, piecewise-constant nature, which prevents gradient-based learning. To overcome this, a line of research has focused on developing a continuous and differentiable surrogate, which we denote as a differentiable Top-K operator $f_{k}(x)$. We begin by formalizing the definition of a standard differentiable Top-K operator. Subsequently, we will discuss several desirable mathematical properties that a well-formulated differentiable relaxation of this operator should possess.

First, we define the set: 
\begin{equation*}
\adjustbox{width=\linewidth}{
$A = \Bigl\{f_{k}(x) =(f_{k}(x_{1}),f_{k}(x_{2}),\dots ,f_{k}(x_{n}))\Bigm|f_{k}(x_{i})\in[0,1],\;\sum_{i=1}^{N} f_{k}(x_{i})=k\Bigr\}$
}    
\end{equation*}
A well-formulated differentiable Top-K operator should map the input vector $x\in \mathbb{R}^{n}$ to a target set $A$, while adhering to the following key properties as closely as possible:
\begin{enumerate}[label=\arabic*), left=0pt]
\item \textbf{Monotonicity.} The operator should be monotonic with respect to its inputs.
\begin{equation*}
f_k(x_i) \geq f_k(x_j)\leftrightarrow x_i \geq x_j
\end{equation*}
\item \textbf{Translation Invariance.} The selection of the top K items should depend only on the relative differences between $x_i$, not their absolute magnitudes. Formally, for any constant c, the set of selected top-K items should be the same for $x$ and $x+c$. 
\begin{equation*}
f_k(x) = f_k(x+c),\quad \forall c \in \mathbb{R}
\end{equation*}
\item \textbf{Approximation Property.} As a temperature-like parameter $\tau$ approaches zero, the $f_{k}(\frac{x}{\tau})$ should converge to the hard Top-K operator. This ensures that the relaxation can be made arbitrarily close to the true hard operation, providing a clear connection between the differentiable surrogate and the original problem.
\begin{equation*}
\lim_{\tau \to 0^{+}} f_k\!\left(\frac{x}{\tau}\right) = F_k(x)
\end{equation*}
\end{enumerate}
By designing operators that satisfy these properties, we can directly model the Top-K objective. The output of the soft operator is a vector $\hat{M}=f_{k}(x)$ that approximates the sparse result of the hard Top-K selection. The optimization objective can then be formulated as maximizing the sum of scores for ground truth within this output vector:
\begin{equation*}
\mathbb{E}\left\lbrack TopK\right\rbrack = -\sum\limits_{i=1}^{N} y_{i} \cdot \hat{m}_{i}
\end{equation*}
The differentiability of this expectation thus allows us to construct a loss function aimed at its maximization, which in turn facilitates the end-to-end optimization of the Top-K problem.

Although constructing a monotonic and differentiable function $f(x_{i})\in[0,1]$ is straightforward (as any cumulative distribution function (CDF) satisfies these conditions), it is non-trivial to simultaneously ensure that the global constraint $\sum_{i=1}^{n} f(x_{i})=k$ is satisfied for all inputs vector $x$.  To enforce this global cardinality constraint, we introduce a corresponding adaptive threshold, denoted as $\theta(x)$, which is a function of the entire input vector $x$. This scalar threshold dynamically shifts the input scores. Specifically, for any given input $x$, we find a unique value for $\theta(x)$ such that the following condition is met:
$$\sum_{i=1}^{N} f(x_i - \theta(x)) = k$$
By solving for this threshold, we ensure the sum-to-$K$ property holds, making $f_k(x) =f(x - \theta(x))$ a strictly differentiable Top-K operator. LapSum~\cite{struski2025lapsum} constructs a specially designed CDF that admits a closed-form solution for all inputs. However, it requires explicit sorting of the inputs before computation, leading to a time complexity of $O(n\log n)$. 
\textbf{Although the existing differentiable Top-K methods have a time complexity of $O(n\log n)$, which is an improvement over the $O(n^2)$ complexity of differentiable sorting methods, their computational complexity remains sub-optimal, which may pose limitations for large-scale data applications in industrial scenarios.} Therefore, there is an urgent need in the industry for an operator with low computational complexity and strong performance.

\subsection{DFTopK: A Fast Differentiable Top-K Operator}
\label{dftopk}
From the perspective of Top-K operator design, based on the principles of translation invariance and approximation property, we can derive that $\theta(x)$ should satisfy the following properties. 1) To ensure approximation property, the value of $\theta(x)$ should lie with in the interval $(x_{[k]},x_{[k+1]})$, where $x_{[k]}$ and $x_{[k+1]}$ denote the $k^{th}$ and ${(k+1)}^{th}$ largest values in the input vector $x$, respectively. 2) To ensure translation invariance, the $\theta(x)$ must satisfy $\theta(x+c)=\theta(x)+c$. Considering time complexity, an intutive solution is $\theta(x)=\alpha x_{[k]}+(1-\alpha)x_{[k+1]},\alpha \in(0,1)$. Ultimately, we choose $\alpha = \frac{1}{2}$. 
For any Top-K function $f_k(x)$, we construct the following unified closed-form solution:
$$\theta(x)=\frac{x_{[k]}+x_{[k+1]}}{2}$$
 The $k^{th}$ largest element in a sequence of length n can be found in average case O(n) linear time using a selection algorithm like \textit{Introselect}. To ensure that our operator and its corresponding loss function yield well-behaved gradients, we specifically select the sigmoid function to serve as the core of our Top-K function.
$$\sigma(x)=\frac{1}{1+e^{-x}}$$
$$f_k(x)=\sigma(x-\theta(x))=\frac{1}{1+e^{-(x-\frac{x_{[k]}+x_{[k+1]}}{2})}}$$
Next, we will prove that our proposed Top-K operator satisfies the three properties mentioned in the previous subsection.
\begin{enumerate}[label=\arabic*),left=0pt] 
\item \textbf{Monotonicity.} The monotonicity of our operator is a direct consequence of the monotonically increasing nature of the sigmoid function.

\item \textbf{Translation Invariance.} By incorporating the offset constant into our equation, we obtain:
\begin{equation*}
f_k(x+c)=\sigma((x+c)-\frac{x_{[k]}+c+x_{[k+1]}+c}{2})=f_k(x)
\end{equation*}
\item \textbf{Approximation Property.} We now proceed to formally prove the Approximation Property. Our goal is to demonstrate that as the temperature parameter $\tau$ approaches zero, our soft operator $f_k(\frac{x}{\tau})$ converges point-wise to the hard Top-K operator $F_{k}\left(x\right)$.
\begin{equation*}
    \begin{aligned}
    \lim_{\tau\to0^{+}}f_{k}\!\left(\frac{x}{\tau}\right)
    &=\sigma\left(\frac{x-\frac{x_{[k]}+x_{[k+1]}}{2}}{\tau}\right)
    =\begin{cases}
    0, & x-\frac{x_{[k]}+x_{[k+1]}}{2}<0 \\
    1, & x-\frac{x_{[k]}+x_{[k+1]}}{2}>0
    \end{cases} \\
    &=\begin{cases}
    0, & x<\frac{x_{[k]}+x_{[k+1]}}{2}\leqslant x_{[k]} \\
    1, & x>\frac{x_{[k]}+x_{[k+1]}}{2}\geqslant x_{[k+1]}
    \end{cases} \\
    &=F_{k}\left(x\right)
    \end{aligned}
\end{equation*}
\end{enumerate}

While our operator relaxes the strict constraint $\sum_{i=1}^{N} f_{k}(x_{i})=k$, the temperature parameter $\tau$ serves as a crucial control knob. As dictated by the approximation property, by tuning $\tau$, we can make our operator's behavior arbitrarily close to that of the hard operator, ensuring that the sum of its outputs $\sum_{i=1}^{N} f_{k}(x_{i})$, converges to a value within a tight neighborhood of $K$.

To optimize our Top-K objective, we formulate a loss function based on Binary Cross-Entropy (BCE). Let $y\in\{0,1\}^N$ be the ground truth vector indicating the Top-K items, and let $f_k(x)$ be the output of our differentiable Top-K operator, where each component $f_k(x_i)$ represents a soft selection probability. The loss is then defined as:
$$\mathcal{L}_{TopK} =-\frac{1}{N}\sum_{i=1}^{N} y_{i}\log f_k(x_i)+(1-y_i)\log (1-f_k(x_i))$$
This formulation directly encourages the model to assign higher probabilities ($f_k(x_i)\longrightarrow 1$) to ground truth items ($y_i=1$) and lower probabilities ($f_k(x_i)\longrightarrow 0$) to others, thereby effectively optimizing for the Top-K selection goal.
It is crucial to clarify why we apply this loss to the output of a listwise Top-K operator $f_k(x)$, instead of treating the problem as a simple pointwise binary classification without an intermediate operator. A pointwise approach, which would directly apply BCE loss to the model's raw logits, such as sigmoid($x$), incorrectly assumes that samples are independent and identically distributed (i.i.d.). This assumption is fundamentally violated in ranking, as an item's Top-K status is highly context-dependent on other competing items within the same page view (PV). Such a mismatch forces a pointwise model to learn from conflicting labels for the same user-item pair across different PVs.
Our listwise operator $f_k(x)$, in contrast, computes a context-aware selection probability 
$f_k(x_i)$ by considering the item's score relative to others in the list. By applying the BCE loss to these context-aware probabilities, we correctly model the conditional nature of the Top-K problem and enable a robust and effective optimization.

We now show that our method enables the backpropagation of effective gradients to the input $x$. By applying the chain rule to our proposed Top-K operator and loss function, we can obtain the gradient of the loss $\mathcal{L}_{TopK}$ with respect to the input vector $x$.
$$\frac{\partial\mathcal{L}_{TopK}}{\partial x} = \frac{\partial\mathcal{L}_{TopK}}{\partial \sigma(x')} \cdot \frac{\partial \sigma(x')}{\partial x'} \cdot \frac{\partial x'}{\partial x}$$
$$x'=x-\frac{x_{[k]}+x_{[k+1]}}{2}$$
where the $\sigma(x)$ is the sigmoid function.

Applying the chain rule, we obtain the following gradient value:
\begin{flalign*}
& \frac{\partial\mathcal{L}_{TopK}}{\partial x_i} = -(y_{i}(1-\sigma(x_{i}^{\prime}))+(1-y_{i})\sigma(x_{i}^{\prime})) & \\
& \qquad = \begin{cases} 
                \sigma(x_i')-1 < 0 & \text{if } y_i = 1 \\
                \sigma(x_i') > 0 & \text{if } y_i = 0
              \end{cases}
             , \quad \forall i\notin\{k,k+1\} &
\end{flalign*}

\begin{equation*}
    \begin{aligned}
        \frac{\partial\mathcal{L}_{TopK}}{\partial x_i} &= -(y_{i}(1-\sigma(x_{i}^{\prime}))+(1-y_{i})\sigma(x_{i}^{\prime})) \\
              &\quad + \frac{1}{2}\sum_{i=1}^{n}(y_{i}(1-\sigma(x_{i}^{\prime}))+(1-y_{i})\sigma(x_{i}^{\prime})),
                \quad\forall i\in\{k,k+1\}
    \end{aligned}
\end{equation*}
The detailed proof can be found in Appendix~\ref{app:proof}. Crucially, our operator's gradient is deterministic for all but two items: $k^{th}$ and $(k+1)^{th}$, which form the decision boundary. This localization of uncertainty to a mere two dimensions drastically mitigates the gradient conflicts common in other methods. Consequently, our model's optimization is stable across a vast majority of dimensions, effectively converging to a neighborhood of a local minimum while only allowing for minor fluctuations within a tiny subspace. Furthermore, by employing a temperature-scaled function $f_{k}(\frac{x}{\tau})$, the temperature parameter $\tau$ itself serves as an explicit mechanism to control the size of the neighborhood around the local minimum to which the model converges.

\section{Experiments}
One key motivation for developing DFTopK is to design a method that offers a closed-form solution with reduced computational complexity while achieving comparable or superior performance. We conducted extensive evaluations on public and industrial datasets, accompanied by detailed analyses, to verify the core effectiveness of our approach. The results demonstrate that DFTopK not only significantly shortens the training time and alleviates gradient conflicts but also reveals its potential to scale effectively with larger datasets. In this work, we mainly present the setup for public experiments, with detailed descriptions of online experiments provided in Section~\ref{sec:online-experiments}.
\subsection{Offline Experiment Setup}
\noindent\textbf{Public Benchmark.} For our public evaluation, we utilize the Rec-Flow benchmark~\cite{liu2024recflow}, which is the only publicly available dataset comprising data from all stages of an operational cascade ranking system. To ensure a comprehensive and fair comparison, we benchmark a suite of leading differentiable Top-K and ranking methods within the experimental framework established by LCRON~\cite{wang2025learning}. We leverage the official open-source implementation of LCRON, guaranteeing that our evaluation utilizes the identical dataset, model architecture, and loss function. Consequently, the sole variable in our comparative study is the differentiable Top-K and ranking operator itself.

\noindent\textbf{Evaluation Protocols.} To construct a realistic two-stage cascade training process, we employ the five designated sample types provided by the benchmark: $rank\_pos$, $rank\_neg$, $coarse\_neg$, $prerank$ $\_neg$, and $sampling\_neg$. Our primary goal is to evaluate the end-to-end performance of the cascade ranking system. To this end, we adopt $Recall@k@m$, as defined in Section~\ref{sec:problem formulation}, as our golden metric. This metric holistically measures the system's ability to retrieve ground truth items(k) from an intermediate candidate set(m). For all our public experiments, we set m=20 and k=10.

However, a critical mismatch exists between public benchmarks and real-world systems, as the limited candidate set size per query in public datasets fails to reflect the scale of industrial applications. To address this gap and rigorously evaluate the scalability of different operators with respect to data volume, we extend the candidate set for each query by uniformly adding $N$ random negatives, thereby constructing a more challenging testbed. In our original dataset, each PV contains 40 samples. \textbf{In the test set, we set $N=160$, resulting in a PV length of 200. This setting not only reflects the larger candidate set sizes in industrial scenarios but also ensures the computational tractability of the experiments with limited GPU resources.} In addition, we further design experiments in Section ~\ref{depth analysis} where the number of training samples per PV is progressively increased. This experimental protocol aims to simulate real-world scenarios, making the experimental results on open-source datasets more convincing.
 
\noindent\textbf{Baselines.}
The operators under evaluation are categorized into two main families: differentiable sorting methods (SoftSort~\cite{prillo2020softsort}, NeuralSort~\cite{grover2019stochastic}, DiffSort~\cite{petersen2022monotonic}) and direct differentiable Top-K operators (Sparse Top-K~\cite{sander2023fast}, Lapsum~\cite{struski2025lapsum}, and our proposed DFTopK). Here, we do not compare with other baseline methods in LCRON~\cite{wang2025learning}, as its approach of modeling the Top-K objective using the NeuralSort operator has already achieved state-of-the-art (SOTA) performance.

\noindent\textbf{Implementation Details.} All experiments were conducted on a single NVIDIA A800 GPU equipped with 80GB of memory, paired with an Intel Xeon Platinum 8352Y CPU. This hardware configuration provides sufficient computational capacity to handle both large-scale training data and complex model architectures in a stable and efficient manner. All offline experiments were implemented in Python 3.7, with PyTorch 1.13 serving as the deep learning framework. For model optimization, we employed the Adam optimizer~\cite{kingma2014adam} with a learning rate of 0.01 and a batch size of 1024 across all methods, ensuring a consistent training environment for fair comparison. Following common practice in online recommendation systems, each model was trained for one epoch on streaming data, which allows us to simulate real-world industrial settings where models are updated continuously with incoming user interactions. This setup not only ensures reproducibility but also enables us to rigorously evaluate the efficiency and effectiveness of different operators under practical constraints. The source code of our public experiments is publicly available\footnote{https://github.com/zhangzhen97/DFTOPK}.

\subsection{Main Results}
\label{main result}
\noindent\textbf{Performance Analysis.} To ensure fairness and industrial relevance, we designed and adopted a rigorous experimental protocol. During training, following the settings in LCRON~\cite{wang2025learning}, we fixed the Top-K parameter K to 10 and set the PV length to 40 across all compared methods, thereby ensuring consistency in optimization objectives. In the testing phase, to better simulate the larger candidate set sizes typically observed in industrial scenarios while keeping the computational cost tractable, we uniformly set the sequence length of all PVs to 200, thus establishing a more challenging evaluation environment. We set the temperature parameter $\tau$ in DFTopK to 500. Table~\ref{tab:main_results} presents the main experimental results on the public RecFlow dataset~\cite{liu2024recflow}, which enables a comprehensive evaluation of different methods under the end-to-end joint Recall metric. 
Our method achieves state-of-the-art performance on this metric, demonstrating its effectiveness in addressing the Top-K issue. As discussed in Section~\ref{intro}, this result also confirms that our operator effectively mitigates the gradient conflict problem present in differentiable sorting methods. Notably, although DFTopK does not outperform all baselines across every single-stage metric (e.g., the Recall of the ranking model), it delivers substantial gains in joint metrics, underscoring its advantages for cascade systems and aligning well with their core objectives.

\begin{table}[!t]
\centering
\caption{Main results of public experiments on RecFlow. To ensure reproducibility, each method was executed with a fixed random seed. Bold numbers indicate the best performance for each metric. Notably, joint Recall@10@20 is regarded as the core metric for evaluating the overall cascade ranking system.}
\label{tab:main_results}
\resizebox{\linewidth}{!}{
\begin{tabular}{lccc}
\toprule
\textbf{Method} &
\makecell[c]{\textbf{Joint}\\\textbf{Recall@10@20}} &
\makecell[c]{\textbf{Ranking}\\\textbf{Recall@10@20}} &
\makecell[c]{\textbf{Retrieval}\\\textbf{Recall@10@30}} \\
\midrule
NeuralSort  & 0.3815 & \textbf{0.4124} & 0.4542 \\
SoftSort    & 0.3988 & 0.3986 & 0.4978 \\
DiffSort    & 0.2465 & 0.2381 & 0.3122 \\
LapSum      & 0.3922 & 0.4043 & 0.4835 \\
Sparse Top-K & 0.3437 & 0.3593 & 0.4299 \\
DFTopK (Ours) & \textbf{0.4040} & 0.4007 & \textbf{0.5069} \\
\bottomrule
\end{tabular}
}
\end{table}

\noindent\textbf{Runtime Analysis:} In this section, we conduct a rigorous empirical analysis to validate the computational efficiency and scalability of our proposed DFTopK operator. Our evaluation is twofold: a direct runtime comparison across different hardware platforms and a scalability analysis with respect to the input dimension N.
First, we benchmark the wall-clock time of a single, complete training iteration (one forward and one backward pass) for DFTopK against all baseline methods. Following the experimental settings in Lapsum~\cite{struski2025lapsum}, to provide a comprehensive evaluation, we conducted experiments on both a standard CPU and a high-performance GPU. This allows us not only to compare the raw speed of each operator but also to assess their efficiency and compatibility with different hardware architectures.
Second, to specifically evaluate scalability, we designed an experiment to measure how the runtime scales with the input dimension N. We varied the sequence length N over a wide range from 5 to 1000, a typical range for many recommendation scenarios. In each configuration, the number of selected items K was kept proportional to the input size by setting $K=\lfloor N/2 \rfloor$
. The superior efficiency of our approach is evidenced by the GPU runtime results presented in Table~\ref{tab:gpu_runtime}. Additional CPU runtime results are included in Appendix~\ref{CPU} for comprehensive analysis. Across both hardware environments, DFTopK consistently achieves the lowest runtime compared to all baselines. More importantly, this advantage becomes increasingly pronounced as the dimension N grows. This empirical finding aligns perfectly with our theoretical complexity analysis, confirming that our $O(n)$ operator successfully breaks the super-linear scaling barrier of $O(n\log n)$ and $O(n^2)$ methods. This linear-time performance is a critical feature, establishing DFTopK as a viable and highly practical solution for real-world systems where efficiently scaling up training samples is a key requirement.
\begin{table}[t]
\caption{GPU Runtime Comparison for a single forward-backward pass. The table reports the average wall-clock time (in ms) for a single forward and backward pass of our proposed DFTopK, differentiable sorting methods, and other Top-K operators. We evaluate the scalability by varying the sequence length $N$ and setting $K=\lfloor N/2 \rfloor$.}
\label{tab:gpu_runtime}
\centering
\resizebox{\columnwidth}{!}{%
\begin{tabular}{c|cccccc} 
\hline
\multirow{2}{*}{Method} & \multicolumn{6}{c}{Runtime (ms)} \\ \cline{2-7}       
                        & N=5 & N=10 & N=50 & N=100 & N=500 & N=1000 \\ \hline
NeuralSort              & 5.57 & 4.87 & 4.96  & 5.14  & 5.78  & 5.42  \\
SoftSort                & 4.78 & 4.34 & 4.57  & 4.34  & 4.96  & 4.34   \\
DiffSort                & 12.12 & 19.11 & 72.88  & 167.38  & 4308.32  & 13781.98  \\
LapSum                  & 6.35 & 5.80 & 6.45  & 5.57  & 7.21   & 5.97  \\
Sparse Top-K            & 549.79 & 533.67 & 552.65  & 520.48  &594.07 & 516.67  \\
DFTopK (ours)       & \textbf{3.55} & \textbf{3.76} & \textbf{3.64} & \textbf{3.47}  & \textbf{4.18}  & \textbf{3.90}  \\ \hline
\end{tabular}
}
\end{table}

\subsection{In-depth Analysis}
\label{depth analysis}
\begin{figure}[!t]
  \centering
  \begin{minipage}{\columnwidth}
    \centering
    \includegraphics[
        width=0.65\linewidth
    ]{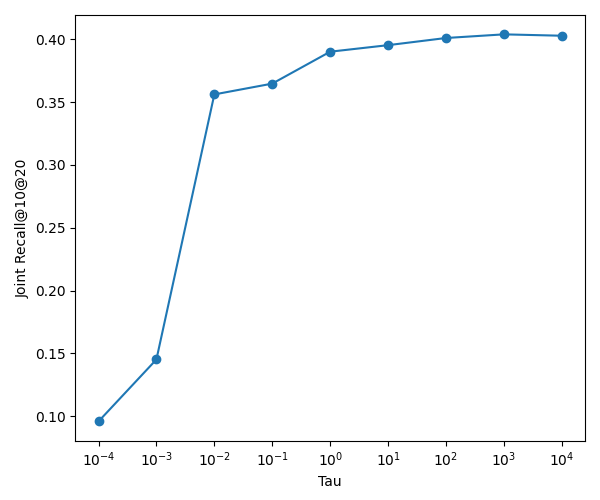}
    \vspace{-0.5cm}
    \caption{Sensitivity Analysis of $\tau$. This figure shows the effect of $\tau$ on our operator. It illustrates the trade-off between approximation hardness and gradient magnitude, demonstrating robustness across a reasonable range.}
    \label{sensitivity}
    \vspace{-0.2cm}
    \label{fig1}
  \end{minipage}
\end{figure}

\noindent\textbf{Sensitivity Analysis.}  Figure ~\ref{sensitivity} illustrates the impact of the adjustable parameter $\tau$ on the behavior of the operator. When $\tau$ takes smaller values, the operator approaches a hard Top-K function, leading to smaller gradients during model updates. In contrast, larger values of $\tau$ yield a smoother operator, accompanied by larger gradients. The experimental results demonstrate that the method exhibits strong robustness as long as $\tau$ remains within a non-extreme range. This indicates that the proposed approach shows low sensitivity to $\tau$, thereby ensuring stability and reliability in practical applications. More detailed numerical results are provided in the appendix~\ref{Sensitivity data}.

\begin{figure}[!t]
  \centering
  \includegraphics[
      width=0.9\linewidth
  ]{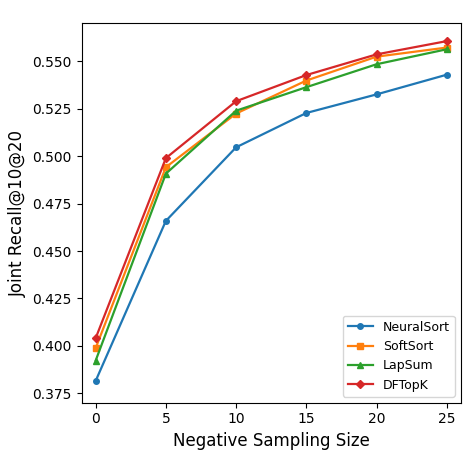}
  \vspace{-0.2cm}
  \caption{Performance under varying negative sampling sizes (Top-4 methods shown). DFTopK consistently achieves SOTA performance across multiple data-scaling settings, demonstrating its efficiency and robustness.}
  \label{fig:data_scaling}
\end{figure}

\noindent\textbf{Data Scaling Performance.} In Section ~\ref{main result}, we empirically demonstrate the runtime advantage of our method on GPU, while Section ~\ref{dftopk} theoretically verifies its time complexity. The linear time complexity property of our operator facilitates performance improvements in industrial scenarios by enabling efficient sample expansion. To further validate the model’s adaptability in practical settings and demonstrate the strong potential of DFTopK for data scaling, we conducted experiments comparing performance under varying numbers of negative samples. As shown in Figure~\ref{fig:data_scaling}, where the Top-4 methods are displayed for clarity, DFTopK consistently achieves state-of-the-art performance across multiple data-scaling settings (see Appendix~\ref{data Scaling up} for complete results). These results confirm the efficiency and stability of DFTopK in recommendation tasks, highlighting its capability to effectively address the increasing scale and complexity of modern recommendation systems.

\begin{figure}[h]
  \centering
  \includegraphics[
      width=0.9\linewidth
  ]{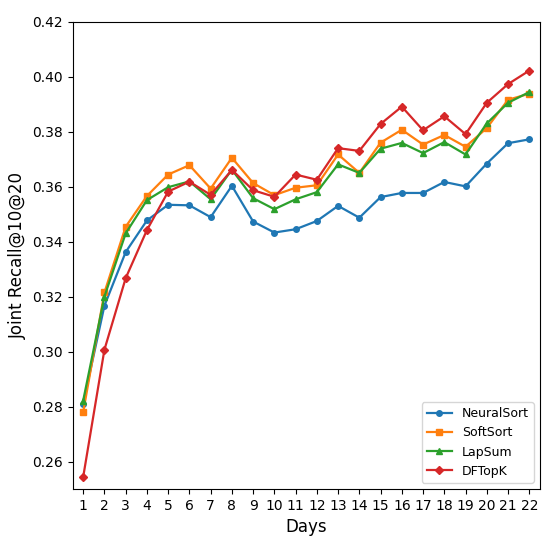}
  \vspace{-0.2cm}
  \caption{Streaming evaluation of Top-4 methods. DFTopK shows superior adaptability and long-term stability across dynamic data streams.}
  \label{fig:streaming}
  \vspace{-0.2cm}
\end{figure}

\noindent\textbf{Streaming Evaluation Analysis.} We further conduct a streaming evaluation to assess model performance under realistic online training conditions. The evaluation protocol is designed as follows: each day's data is treated as an independent test set, while all preceding historical data are used for training. Figure~\ref{fig:streaming} presents the training curves of the Top-4 performing methods. As demonstrated, in the very early stages of training, our method temporarily lags behind some baselines. However, as more training data is accumulated over time, our model exhibits superior learning efficiency. Around day 10, DFTopK begins to consistently outperform all competing methods and continues to widen its performance gap thereafter. These results confirm the strong adaptability and robustness of our approach in dynamic, real-world industrial environments. More importantly, they highlight its superior long-term learning potential.

\subsection{Online Deployment}
\label{sec:online-experiments}
While DFTopK demonstrates consistent improvements across different settings on the RecFlow benchmark, a reported gain there does not directly indicate how much it can improve key business metrics—such as revenue or conversion—in a real-world advertising system. To evaluate its practical impact, online deployment and A/B testing in a large-scale industrial environment are essential.

We deployed the DFTopK operator, integrated within the LCRON framework, to the Retrieval and Pre-ranking stages of a real-world advertising system. Due to the space limitation, implementation details are described in Appendix~\ref{app:online}. For the online A/B test, constrained by computational resources, we selected a single, strong, and representative baseline for comparison. Specifically, our online baseline is NeuralSort, a method that has demonstrated proven effectiveness on industrial benchmarks. All experimental groups were trained for seven days using an online learning paradigm before being evaluated over a 15-day online testing period.

The results of our online A/B test are detailed in Table~\ref{ABtest}. First, under an isosample condition (i.e., with an identical number of training samples), DFTopK achieves a 0.57\% improvement in advertising revenue and a 0.9\% improvement in advertising conversions compared to the NeuralSort baseline. Concurrently, in terms of efficiency, our method significantly reduces the average rungraph time per impression from 150\textmu s to 127\textmu s, yielding a 15.3\% fold speedup.

Furthermore, by reducing training costs, it enables efficient data scaling, thereby bringing tangible benefits. To validate this, we scaled up the number of negative samples in DFTopK such that its training runtime matched that of the NeuralSort baseline. Experimental results show that with the increased negative sampling, DFTopK achieves further substantial gains, achieving a significant 1.77\% improvement in online revenue and a 1.5\% improvement in conversions compared to the baseline.

\begin{table}[H]
  \caption{Industrial experimental results for 15 days on a real-world advertising system. Each method was allocated 10\% of the traffic, and online metrics were evaluated using relative improvement, with NeuralSort serving as the baseline for comparison. We scaled up the number of negative samples in DFTopK so that its training runtime matched that of the NeuralSort baseline.}
  \label{ABtest}
  \label{tab:online_exp}
  \centering	
  \resizebox{1.0\linewidth}{!}{
  \begin{tabular}{c|c|c|c}
    \hline
    \multirow{2}{*}{Method/Metric} & \multicolumn{3}{c}{Online Metrics}\\ 
    \cline{2-4}
    & Revenue & Ad Conversions & \makecell[c]{RunGraph Time \\ Per Impression} \\ 
    \hline
    NeuralSort & -- & -- & 150\textmu s\\
    DFTopK & +0.57\% & +0.9\% & 127\textmu s (-15.3\%)\\  
    DFTopK+Data Scaling & +1.77\% & +1.5\% & 150\textmu s\\ 
    \hline
  \end{tabular}
  }
\end{table}
\section{Conclusion}

This work proposes a novel differentiable Top-K operator, DFTopK, which achieves a balance between the desirable properties of strictly differentiable operators and computational efficiency. By relaxing the normalization constraint and introducing a temperature-controlled approximation mechanism, DFTopK attains $O(n)$ linear-time complexity while preserving stable and well-behaved gradient characteristics. Theoretical analysis further shows that DFTopK effectively mitigates the gradient conflicts inherent in differentiable sorting methods for the Top-K selection problem. Extensive experiments on public datasets and industrial systems demonstrate that DFTopK significantly improves efficiency while maintaining strong performance, offering a fast, stable, and theoretically sound solution for Top-K modeling in real-world applications.

\begin{acks}
The authors would like to express their sincere thanks to Zhiliang Zhu, Chao Wang, and Yanhua Cheng for their insightful discussions on differentiable sorting and Top-K operators. We also gratefully acknowledge Jianlin Su, whose technical blog posts provided valuable inspiration for our work on differentiable Top-K problems.
\end{acks}

\bibliographystyle{ACM-Reference-Format}
\bibliography{sample-base}

\newpage
\onecolumn 
\appendix

\section{Gradient Derivation of the Loss $\mathcal{L}_{TopK}$ w.r.t. Input $x$ }\label{app:proof}
The loss function is defined as follows:
$$\mathcal{L}_{TopK} =-\frac{1}{N}\sum_{i=1}^{N} y_{i}\log f_k(x_i)+(1-y_i)\log (1-f_k(x_i))$$
By applying the chain rule to our proposed Top-K operator and loss function, we can obtain the gradient of the loss $\mathcal{L}_{TopK}$ with respect to the input vector $x$.
$$\frac{\partial\mathcal{L}_{TopK}}{\partial x} = \frac{\partial\mathcal{L}_{TopK}}{\partial \sigma(x')} \cdot \frac{\partial \sigma(x')}{\partial x'} \cdot \frac{\partial x'}{\partial x}$$
where $x'$ represents the translated input vector, defined as:
$$x'=x-\frac{x_{[k]}+x_{[k+1]}}{2}$$
\noindent\textbf{Derivative of $\mathcal{L}_{TopK}$ with respect to $\sigma(x')$.} By differentiating the natural logarithm function, we can readily obtain:
\begin{equation*}
    \begin{aligned}
    \frac{\partial\mathcal{L}_{TopK}}{\partial \sigma(x')}
    & = \left[\frac{\partial\mathcal{L}_{TopK}}{\partial \sigma(x_1')},\cdots,
            \frac{\partial\mathcal{L}_{TopK}}{\partial \sigma(x_N')}\right] \\
    & = \left[-\left(\frac{y_i}{f_k(x_i)}+\frac{1-y_i}{1-f_k(x_i)}\right),\cdots,
            -\left(\frac{y_{N}}{f_k(x_N)}+\frac{1-y_N}{1-f_k(x_N)}\right)\right]
    \end{aligned}
\end{equation*}
\noindent\textbf{Derivative of $\sigma(x')$ with respect to $x'$.} From the derivative of the sigmoid function, we can directly obtain:
$$\frac{\partial \sigma(x')}{\partial x'} =diag(\sigma(x_1')(1-\sigma(x_1')),\cdots,\sigma(x_N')(1-\sigma(x_N')))$$
\noindent\textbf{Derivative of $x'$ with respect to $x$.} Due to $x'=x-\frac{x_{[k]}+x_{[k+1]}}{2}$, we can obtain:
$$\frac{\partial x'}{\partial x} = \mathbb{I}_N + \begin{bmatrix}
  &0  &\cdots  &-\frac{1}{2} &-\frac{1}{2} &\cdots &0   \\
  &\vdots  &\vdots  &\vdots &\vdots &\vdots &\vdots    \\
  &0  &\cdots  &-\frac{1}{2} &-\frac{1}{2} &\cdots &0   \\
\end{bmatrix}$$
where $\mathbb{I}_N$ is the $N\times N$ identity matrix, and the other is a matrix of the same dimensions whose entries are all zero, except for the $k^{th}$ and $(k+1)^{th}$ columns where all entries are $-\frac{1}{2}$. By applying the chain rule, we obtain the following gradient value:

\begin{align*}
\frac{\partial\mathcal{L}_{TopK}}{\partial x_i} 
&= -(y_{i}(1-\sigma(x_{i}^{\prime}))+(1-y_{i})\sigma(x_{i}^{\prime}))= \begin{cases} 
    \sigma(x_i')-1 < 0 & \text{if } y_i = 1 \\
    \sigma(x_i') > 0 & \text{if } y_i = 0
   \end{cases},
\quad \forall i\notin\{k,k+1\} \\
\frac{\partial\mathcal{L}_{TopK}}{\partial x_i} &= -(y_{i}(1-\sigma(x_{i}^{\prime}))+(1-y_{i})\sigma(x_{i}^{\prime})) + \frac{1}{2}\sum_{i=1}^{N}(y_{i}(1-\sigma(x_{i}^{\prime}))+(1-y_{i})\sigma(x_{i}^{\prime})),
\quad \forall i\in\{k,k+1\}
\end{align*}

Our operator exhibits deterministic gradients for all elements except the $k^{th}$ and $(k+1)^{th}$ positions, which together determine the decision boundary.
By concentrating stochasticity within this narrow two-dimensional region, the method effectively avoids the widespread gradient conflict problem observed in many differentiable sorting methods. This design ensures that the optimization dynamics remain well-behaved across most dimensions, enabling the model to settle smoothly into a stable region around a local optimum, with only minimal variation in a very limited subspace.
In addition, the temperature parameter $\tau$ in the scaling function $f_k(\frac{x}{\tau})$ provides a direct control mechanism over the extent of this stability region, serving as a smoothness regulator in the learning landscape.

\section{Runtime of DFTopK on CPU}\label{CPU}
While the main text focuses on GPU performance, we provide comprehensive CPU runtime results in this appendix to offer a complete evaluation of computational efficiency across different hardware architectures. As shown in Table~\ref{tab:cpu_runtime}, DFTopK demonstrates consistent efficiency advantages on CPU platforms, achieving the fastest runtime across all input dimensions. The linear time complexity of our operator translates to even more significant speedups on CPU compared to the $O(n\log n)$ and $O(n^2)$ baselines. These results reinforce the hardware-agnostic efficiency of our approach and highlight its practical applicability in resource-constrained deployment scenarios.

\begin{table}[h]
\caption{CPU Runtime Comparison for a single forward-backward pass. The table reports the average wall-clock time (in ms) for a single forward and backward pass of our proposed DFTopK, differentiable sorting methods, and other Top-K operators. We evaluate the scalability by varying the sequence length $N$ and setting $K=\lfloor N/2 \rfloor$.}
\label{tab:cpu_runtime}
\centering
\resizebox{0.65\linewidth}{!}{%
\begin{tabular}{c|cccccc} 
\hline
\multirow{2}{*}{Method} & \multicolumn{6}{c}{Runtime (ms)} \\ \cline{2-7}       
                        & N=5 & N=10 & N=50 & N=100 & N=500 & N=1000 \\ \hline
NeuralSort              & 3.69 & 3.15 & 56.38  & 80.57  & 257.55  & 258.79  \\
SoftSort                & 3.13 & 2.95 & 37.82  & 36.85  & 182.92  & 134.30  \\
DiffSort                & 8.36 & 11.56 & 107.73  & 716.46  & 41103.13  & 66308.17  \\
LapSum                  & 4.02 & 3.44 & 33.24  & 38.83   & 72.64  &  86.43  \\
Sparse Top-K            & 707.16 & 771.36 & 823.36  & 767.05  &748.81   & 976.25  \\
DFTopK (ours)       & \textbf{1.86} & \textbf{2.78} & \textbf{23.08}  & \textbf{32.91}  & \textbf{70.34}  & \textbf{69.13}  \\ \hline
\end{tabular}
} 
\end{table}

\section{Sensitivity Analysis Data}\label{Sensitivity data}

We conduct a sensitivity study on DFTopK with respect to its hyperparameter $\tau$, which regulates the degree of smoothness in the differentiable Top-K operator. Smaller values of $\tau$ make the approximation sharper and closer to a discrete selection, whereas larger values result in smoother transitions. Table~\ref{tab:tau_sensitivity} reports the performance under different $\tau$ configurations, showing that the optimal results are obtained when $\tau=1000$. Notably, even when $\tau$ deviates from this setting (e.g., $\tau=1000$ or $\tau=2000$), DFTopK still delivers substantial gains over the baseline methods (see Table~\ref {tab:main_results}), indicating that the model is relatively insensitive to moderate variations of $\tau$.
The empirical trend reveals a single-peaked relationship between $\tau$ and performance: accuracy improves as $\tau$ increases up to a moderate level and then gradually declines beyond that point. This observation suggests that DFTopK achieves a favorable balance between smoothness and discrimination at intermediate $\tau$ values. In practice, this behavior simplifies hyperparameter tuning, as near-optimal performance can be achieved without fine-grained parameter search.

\begin{table}[h]
\centering
\vspace{-0.8cm}
\caption{Hyperparameter sensitivity analysis of DFTopK with respect to temperature parameter $\tau$. Bold numbers indicate the best performance for each metric.}
\label{tab:tau_sensitivity}
\resizebox{0.55\linewidth}{!}{
\begin{tabular}{lccc}
\toprule
\textbf{$\tau$} &
\makecell[c]{\textbf{Joint}\\\textbf{Recall@10@20}} &
\makecell[c]{\textbf{Ranking}\\\textbf{Recall@10@20}} &
\makecell[c]{\textbf{Retrieval}\\\textbf{Recall@10@30}} \\
\midrule
0.0001 & 0.0959 & 0.0915 & 0.1528 \\
0.001  & 0.1452 & 0.3951 & 0.1528 \\
0.01   & 0.3561 & 0.4044 & 0.4174 \\
0.1    & 0.3647 & 0.4063 & 0.4353 \\
1      & 0.3901 & \textbf{0.4101} & 0.4720 \\
10     & 0.3953 & 0.4057 & 0.4869 \\
100    & 0.4010 & 0.4022 & 0.4990 \\
1000   & \textbf{0.4039} & 0.3871 & 0.5090 \\
10000  & 0.4028 & 0.3917 & \textbf{0.5101} \\
\bottomrule
\end{tabular}
}
\vspace{-0.8cm}
\end{table}

\section{Data Scaling Performance}\label{data Scaling up}
This appendix provides the complete set of experimental results to complement the performance analysis in Section~\ref {depth analysis}. To maintain clarity in visual presentation, Figure~\ref {fig:data_scaling} in the main text only displays the Top-4 performing methods on the core metric Joint Recall@10@20. Here, we provide comprehensive results for all methods across all evaluation dimensions. Table~\ref {tab:scaling1} reports the detailed performance of all six methods, including NeuralSort, SoftSort, DiffSort, LapSum, Sparse Top-K, and DFTopK, covering individual stage metrics (ranking recall and retrieval recall) under different negative sampling sizes ($N_{neg}=0, 5, 10, 15, 20, 25$).
The detailed data reveal two key findings: First, although DFTopK may not achieve the highest score on every individual single-stage metric, it consistently delivers the best performance when these components are integrated in a joint evaluation. This is critical for practical industrial deployment, where overall system efficiency outweighs the performance of individual components. Second, our method remains optimal as the negative sampling size increases, demonstrating superior scalability.

\begin{table}[H]
\caption{Main results of public experiments on RecFlow. To ensure reproducibility, each method was executed with a fixed random seed. Bold numbers indicate the best performance for each metric. Notably, joint Recall@10@20 is regarded as the core metric for evaluating the overall cascade ranking system. The test set consists of data from the last day, with batch-wise negative sampling applied to extend the sequence length to 200, while the remaining data is used for training.}
\vskip -0.1in
\label{tab:scaling1}
\centering
\resizebox{\textwidth}{!}{%
\begin{tblr}{
  colspec = {c | *{4}{ccc}}, 
  cell{1}{1} = {r=2}{m},     
  cell{1}{2} = {c=3}{c}, 
  cell{1}{5} = {c=3}{c}, 
  cell{1}{8} = {c=3}{c}, 
  hline{1,3,9} = {-}{solid},
  hline{2} = {2-13}{solid},
  vline{2,5,8,11} = {1-9}{solid},
}
\bfseries Method/Metric 
  & \bfseries $N_{neg}$=0 & &  
  & \bfseries $N_{neg}$=5 & &      
  & \bfseries $N_{neg}$=10 & & \\
 & \makecell{Joint \\ Recall@10@20} 
 & \makecell{Ranking \\ Recall@10@20} 
 & \makecell{Retrieval \\ Recall@10@30} 
 & \makecell{Joint \\ Recall@10@20} 
 & \makecell{Ranking \\ Recall@10@20} 
 & \makecell{Retrieval \\ Recall@10@30} 
 & \makecell{Joint \\ Recall@10@20} 
 & \makecell{Ranking \\ Recall@10@20} 
 & \makecell{Retrieval \\ Recall@10@30} \\
NeuralSort       &0.3815 &\textbf{0.4124} &0.4542 &0.4659 &0.4991 &0.5238 &0.5047 &0.5232 &0.5715  \\
SoftSort         &0.3988 &0.3986 &0.4978 &0.4942 &0.4951 &0.5752 &0.5225 &0.5200 &0.6043  \\
DiffSort         &0.2465 &0.2381 &0.3122 &0.2560 &0.2574 &0.3230 &0.2733 &0.2738 &0.3487  \\
LapSum           &0.3922 &0.4043 &0.4835 &0.4908 &\textbf{0.5042} &0.5609 &0.5240 &\textbf{0.5324} &0.5867  \\
Sparse Top-K     &0.3437 &0.3593 &0.4299 &0.4026 &0.4407 &0.4642 &0.4392 &0.4733 &0.5053  \\
\bfseries DFTopK (Ours) &\textbf{0.4040} &0.4007 &\textbf{0.5069} &\textbf{0.4990} &0.4929 &\textbf{0.5835} &\textbf{0.5290} &0.5171 &\textbf{0.6180} \\
\end{tblr}%
}
\end{table}

\begin{table}[H]
\vspace{-0.3cm}
\label{tab:scaling2}
\centering
\resizebox{\textwidth}{!}{%
\begin{tblr}{
  colspec = {c | *{4}{ccc}}, 
  cell{1}{1} = {r=2}{m},     
  cell{1}{2} = {c=3}{c}, 
  cell{1}{5} = {c=3}{c}, 
  cell{1}{8} = {c=3}{c}, 
  hline{1,3,9} = {-}{solid},
  hline{2} = {2-13}{solid},
  vline{2,5,8,11} = {1-9}{solid},
}
\bfseries Method/Metric 
  & \bfseries $N_{neg}$=15 & &  
  & \bfseries $N_{neg}$=20 & &      
  & \bfseries $N_{neg}$=25 & & \\
 & \makecell{Joint \\ Recall@10@20} 
 & \makecell{Ranking \\ Recall@10@20} 
 & \makecell{Retrieval \\ Recall@10@30} 
 & \makecell{Joint \\ Recall@10@20} 
 & \makecell{Ranking \\ Recall@10@20} 
 & \makecell{Retrieval \\ Recall@10@30} 
 & \makecell{Joint \\ Recall@10@20} 
 & \makecell{Ranking \\ Recall@10@20} 
 & \makecell{Retrieval \\ Recall@10@30} \\
NeuralSort       &0.5228 &0.5383 &0.5909 &0.5326 &0.5480 &0.6048 &0.5430 &0.5556 &0.6115  \\
SoftSort         &0.5399 &0.5354 &0.6205 &0.5525 &0.5480 &0.6329 &0.5573 &0.5531 &0.6389  \\
DiffSort         &0.2645 &0.2687 &0.3329 &0.1194 &0.1951 &0.1563 &0.2540 &0.2430 &0.3271  \\
LapSum           &0.5364 &\textbf{0.5418} &0.6105 &0.5485 &\textbf{0.5527} &0.6221 &0.5564 &\textbf{0.5570} &0.6330  \\
Sparse Top-K     &0.4516 &0.4877 &0.5186 &0.4504 &0.4938 &0.5156 &0.4547 &0.4960 &0.5231  \\
\bfseries DFTopK (Ours) &\textbf{0.5428} &0.5314 &\textbf{0.6270} &\textbf{0.5537} &0.5414 &\textbf{0.6385} &\textbf{0.5607} &0.5481 &\textbf{0.6470} \\
\end{tblr}%
}
\vskip -0.1in
\end{table}

\section{Implementation Details of Online Experiments}\label{app:online}
This section provides a comprehensive description of our online experimental implementation. While the full production pipeline consists of four consecutive stages (Matching, Pre-ranking, Ranking, and Re-ranking), our experimental setup adopts LCRON's configuration, focusing on a simplified two-stage cascade framework. As shown in Figure~\ref {fig:cascade}, this streamlined architecture maintains the core functionality of complete systems while substantially improving experimental tractability.

\textbf{Model Architectures.} We utilize a Deep Structured Semantic Model (DSSM) for the retrieval phase and a Multi-Layer Perceptron (MLP) for pre-ranking. The DSSM framework implements twin-tower architecture with identical [1024,768,768,96] layer configurations for both user and item encoders. The pre-ranking MLP follows a [1024,768,768,1] dimensional design. The dimension of each sparse feature is set to 96. All hidden layers employ PReLU activation functions, He initialization strategy, and batch normalization with momentum set to 0.999.

\textbf{Training Methodology.} Our DFTopK implementation preserves identical training sample organization with LCRON's online experiments~\cite{wang2025learning}. For data scaling investigations, we scaled up the number of negative samples in DFTopK such that its training runtime matched that of the NeuralSort baseline. The training process utilizes online streaming, processing about 20 billion user-ad pairs daily with AdaGrad optimizer (learning rate 0.01) and a consistent batch size of 4096 for both retrieval and pre-ranking models. All parameters are initialized randomly and trained exclusively within the TensorFlow ecosystem.

The core methodological innovation lies in substituting the NeuralSort operator with our DFTopK operator in the LCRON framework for better end-to-end recall metric optimization. We synchronize the training of retrieval and pre-ranking models in a single training task while producing unified checkpoints. During deployment, we regenerate distinct metadata files for individual models, facilitating separate serving where each model exclusively loads its corresponding parameters from the jointly-trained checkpoint.

\end{document}